\documentclass[10pt,twocolumn,letterpaper]{article}

\usepackage{iccv}
\usepackage[pagebackref=true,breaklinks=true,letterpaper=true,colorlinks,bookmarks=false]{hyperref}
\usepackage{times}
\usepackage{epsfig}
\usepackage{graphicx}
\usepackage{amsmath}
\usepackage{amssymb}
\usepackage{booktabs}
\usepackage{multirow}
\usepackage{bigstrut}
\usepackage{hyperref}
\usepackage{mathrsfs}
\usepackage{rotating}
\usepackage{adjustbox}
\usepackage[table]{xcolor}

\definecolor{my_green}{RGB}{18,159,87}
\usepackage[pagebackref=true,breaklinks=true,letterpaper=true,colorlinks,bookmarks=false]{hyperref}

\newcommand{\ieno}{\textit{i}.\textit{e}.}
\newcommand{\egno}{\textit{e}.\textit{g}.} 

\usepackage{graphicx}
\usepackage{caption}
\usepackage{subcaption}
\usepackage{float}

\iccvfinalcopy 

\def\iccvPaperID{7318} 

\ificcvfinal\fi

\begin{document}

\title{Inpaint Anything: Segment Anything Meets Image Inpainting}






\author{Tao Yu$^{1}$~~Runseng Feng$^{1}$~~Ruoyu Feng$^{1}$~~Jinming Liu$^{2}$~~Xin Jin$^{2}$~~Wenjun Zeng$^{2}$~~Zhibo Chen$^1$\ \\$^1$University of Science and Technology of China ~$^2$Eastern Institute for Advanced Study \\
{\tt\small \{yutao666, fengruns, ustcfry\}@mail.ustc.edu.cn} \\ {\tt\small \{jmliu, jinxin, wenjunzeng\}@eias.ac.cn,} {\tt\small chenzhibo@ustc.edu.cn}}

\ificcvfinal\thispagestyle{empty}\fi

\definecolor{my_green}{RGB}{18,159,87}

\twocolumn[{%

\maketitle

\begin{figure}[H]
\hsize=\textwidth %
\centering
\includegraphics[width=.9\textwidth]{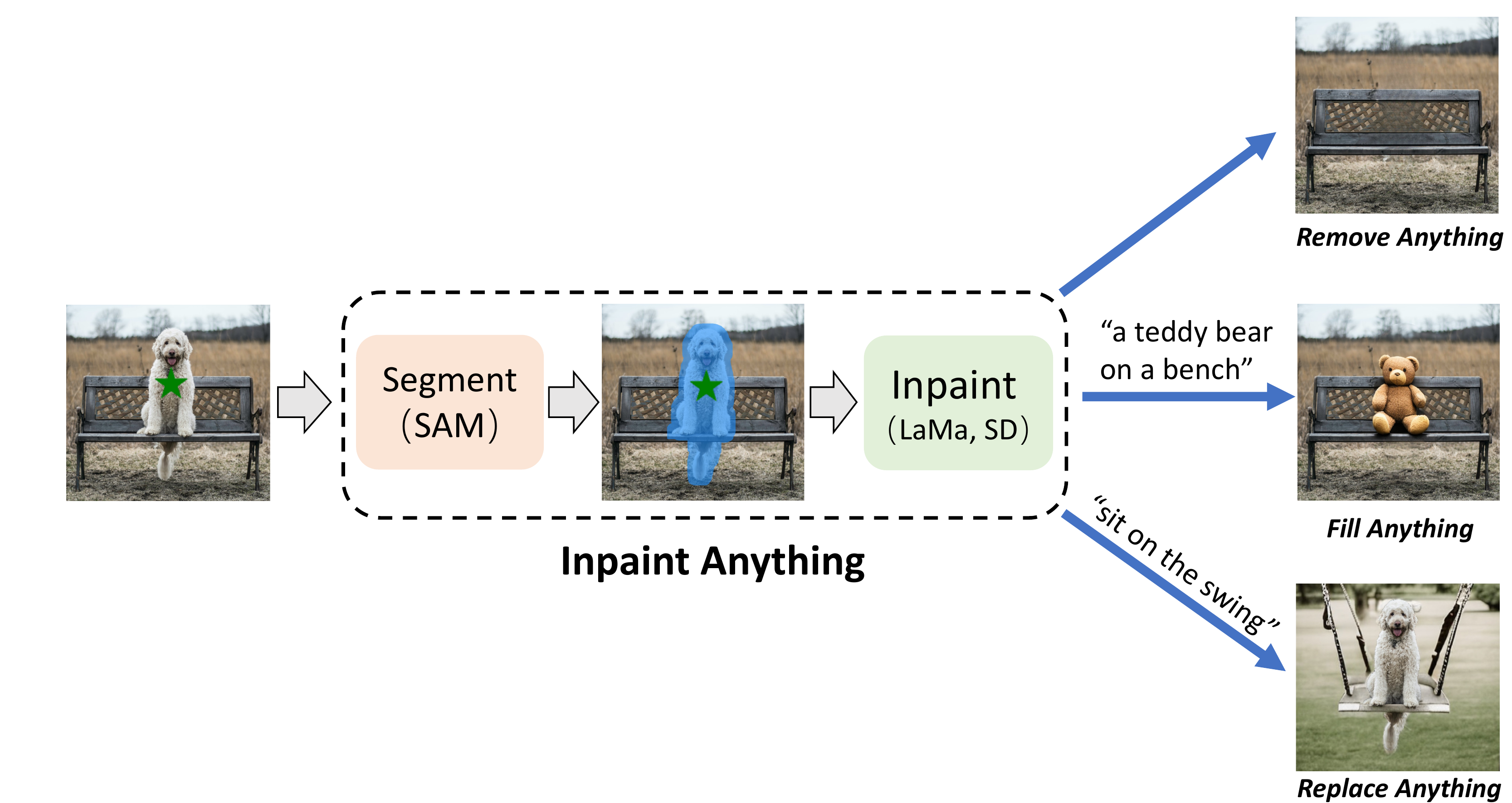}
\caption{Illustration of our Inpaint Anything. Users can select any object in an image by clicking on it. With powerful vision models, \egno, SAM\cite{kirillov2023segment}, LaMa \cite{suvorov2022resolution} and Stable Diffusion (SD) \cite{rombach2022high}, Inpaint Anything is able to remove the object smoothly (\ieno, Remove Anything). Further, by inputting text prompts, users can fill the object with any desired content (\ieno, Fill Anything) or replace the background of it arbitrarily (\ieno, Replace Anything).}
\label{framework}
\end{figure}
}]


\begin{abstract}

Modern image inpainting systems, despite the significant progress, often struggle with mask selection and holes filling. Based on Segment-Anything Model (SAM)~\cite{kirillov2023segment}, we make the first attempt to the mask-free image inpainting and propose a new paradigm of ``clicking and filling'', which is named as Inpaint Anything (IA). The core idea behind IA is to combine the strengths of different models in order to build a very powerful and user-friendly pipeline for solving inpainting-related problems. IA supports three main features: (i) Remove Anything: users could click on an object and IA will remove it and smooth the ``hole'' with the context; (ii) Fill Anything: after certain objects removal, users could provide text-based prompts to IA, and then it will fill the hole with the corresponding generative content via driving AIGC models like Stable Diffusion \cite{rombach2022high}; (iii) Replace Anything: with IA, users have another option to retain the click-selected object and replace the remaining background with the newly generated scenes. We are also very willing to help everyone share and promote new projects based on our Inpaint Anything (IA). Our codes are available at \url{https://github.com/geekyutao/Inpaint-Anything}.

\end{abstract}


\section{Motivation and Observation}

\subsection{Why do we need Inpaint Anything?}

\begin{itemize}

\item The state-of-the-art (SOTA) image inpainting works, like LaMa~\cite{suvorov2022resolution}, Repaint~\cite{lugmayr2022repaint}, MAT~\cite{li2022mat},  ZITS~\cite{dong2022incremental}, \textit{etc}., have achieved great progress. They can successfully inpaint large regions and work well with complex repetitive structures, generalizing well to high-resolution images. However, they typically need fine annotations for each mask, which are essential for training and inference.


\item Segment Anything Model (SAM)~\cite{kirillov2023segment} is a strong segmentation foundation model, producing high quality object masks from input prompts such as points or boxes, and it can be used to generate comprehensive and accurate masks for all objects in an image. However, their mask segmentation predictions have not been fully-explored.

\item Besides, the existing inpainting methods can only fill the removed area with the context. AIGC models open up new opportunities for creation, which has the potential to meet massive demand and assists humans to newly generate their wanted content.

\item Therefore, by combining the advantages of SAM~\cite{kirillov2023segment}, the SOTA image inpainters~\cite{suvorov2022resolution}, and AI generated content (AIGC) models~\cite{rombach2022high}, we provide a powerful and user-friendly pipeline for solving more general inpainting-related problems, such as object removal, new content filling, and background replacing.

\end{itemize} 

\subsection{What Inpaint Anything can do?}

\begin{itemize}

\item \textbf{SAM + SOTA inpainters for removing anything}: With IA, users can easily remove specific objects from the interface by simply clicking on them. Furthermore, IA provides an option for users to fill the resulting ``hole'' with contextual data. Oriented at this, we combine the strengths of SAM and some SOTA Inpainters like LaMa. Once manually refined through corrosion and dilation, the mask predictions generated by SAM serve as input for the inpainting models, providing clear indicators for the object areas to be erased and filled.

\item \textbf{SAM + AIGC models for filling or replacing anything}: 

(1) After removing objects, IA provides users the option to fill the resulting ``hole'' either with contextual data or ``new content''. Specifically, a strong AI generated content (AIGC) model like Stable Diffusion~\cite{rombach2022high} is utilized to generate new objects via text prompts. For example, users can use the word of ``dog'' or a sentence of ``a cute dog, sitting on the bench'', to generate a new dog for filling the hole with such newly generated dog.

(2) In addition, users have another option to take IA to retain the click-selected object and replace the remaining background with the newly generated scene. This scene replacement process of IA supports various ways of prompting AIGC models, such as using a different image as visual prompt or using a short caption as text prompt. For example, users can keep the dog in an image but replace the original indoor background with an outdoor one.

\end{itemize} 







\section{Methodology}

\subsection{Preliminary}

\paragraph{Segment Anything Model (SAM).}
The fundamental CV model of Segment Anything~\cite{kirillov2023segment} was released last week, which is a large ViT-based model trained on the large visual corpus (SA-1B). SAM has demonstrated promising segmentation capabilities in various
scenarios and the great potential of the foundation models for computer vision. This is a ground-breaking step toward visual artificial general intelligence, and SAM was once hailed as ``the CV version of ChatGPT''.

\paragraph{SOTA Inpainters.}
Image inpainting, as an ill-posed inverse problem, is widely explored in the field of computer vision and image processing, which intended to replace missing regions of damaged images with visually plausible structure and texture. The success of deep learning has brought new opportunities~\cite{suvorov2022resolution,lugmayr2022repaint,li2022mat,dong2022incremental}, and all these SOTA methods can be categorized from multiple perspectives, e.g., inpainting strategies, network structures, and loss functions. For our Inpaint Anything (IA), we investigated the use of a simple, single-stage approach LaMa~\cite{suvorov2022resolution} for mask-based inpainting, which is arguably good in generating repetitive visual structures by combining fast Fourier convolutions (FFCs)~\cite{chi2020fast}, perceptual loss~\cite{johnson2016perceptual}, and an aggressive training mask generation strategy.

\paragraph{AIGC Models.}
ChatGPT~\footnote{https://chat.openai.com/} and other Generative AI (GAI) techniques all belong to the category of Artificial Intelligence Generated Content (AIGC), which involves the creation of digital content, such as images, music, and natural language, through AI models. It is considered a new type of content creation and has shown to achieve state-of-the-art performance in various content generation~\cite{rombach2022high,saharia2022photorealistic}. For our work of IA, we directly employ a powerful AIGC model of Stable Diffusion~\cite{rombach2022high} to generate the desired content in
the hole based on text-prompting.

\subsection{Inpaint Anything}
The principle of our proposed Inpaint Anything (IA) is to composite off the shelf foundation models to enable the ability of solving extensive image inpainting problems.
By compositing the strengths of various foundation models, IA can generate high-quality inpainted images. 
Specifically, our IA has three schemes, \ieno, \textit{Remove Anything}, \textit{Fill Anything} and \textit{Remove Anything}, which are designed to \textit{remove}, \textit{fill} and \textit{replace} anything, respectively.

\paragraph{Remove Anything.}
Remove Anything focuses on the object removal problem \cite{criminisi2003object,criminisi2004region,elharrouss2020image} by allowing users to eliminate any object from an image while ensuring that the resulting image remains visually plausible.
Remove Anything consists of three steps: clicking, segmenting, and removing, as shown in Figure \ref{framework}. In the first step, users select the object they want to remove from an image by clicking on it. Next, a foundation segmentation model, such as Segment Anything \cite{kirillov2023segment}, is utilized to automatically segment the object based on the click location and create a mask. Finally, a state-of-the-art inpainting model, such as LaMa \cite{suvorov2022resolution}, is used to fill the hole created by the removed object using the mask. Since the object is no longer present in the image, the inpainting model fills the hole with background information.
Note that, for the entire process, users only need to click on the object they want to remove from the image.


\paragraph{Fill Anything.}
Fill Anything allows users to fill any object in an image with any content they want. The tool consists of four steps: clicking, segmenting, text-prompting, and generating.
The first two steps of Fill Anything are the same as in Remove Anything. In the third step, users input a text prompt that indicates what they want to fill the object hole with. Finally, a powerful AIGC model, such as Stable Diffusion \cite{rombach2022high}, is adopted to generate the desired content in the hole based on the text-prompt inpainting model.

\paragraph{Replace Anything.}
Replace Anything is able to replace any object with any background.
The process of Replace Anything is similar to that of Fill Anything, but in this case, the AIGC model is prompted to generate visually consistent background that exists outside the specified object.

\paragraph{Practice.}
Compositing foundation models to solve tasks may encounter problems such as incompatibility or inappropriateness. We should consider intermediate processing for better coordination between models and tasks. 
In this work, for image inpainting scenario, we summarize several good practice for the composition below.

\begin{itemize}

\item \textbf{Dilation matters.} We observe that the segmentation result (\ieno, object mask) of SAM may contain discontinuous and non-smooth boundaries, or holes inside the object region. These issues pose challenges for effectively removing or filling objects. Consequently, we employ a dilation operation to refine the mask.
Further, for filling objects, large masks give AIGC models more space to create, benefiting the ``alignment'' to user purpose. Thus, we adopt a large dilation in Fill Anything.

\item \textbf{Fidelity matters.} Most state-of-the-art AIGC models such as Stable Diffusion require require images to be of a fixed resolution, typically 512 $\times$ 512. Simply resizing images to this resolution may result in a loss of fidelity, which can adversely impact the final inpainted results. Therefore, it is essential to adopt measures that preserve the original image quality, such as utilizing cropping techniques or maintaining the image's aspect ratio when resizing.

\item \textbf{Prompt matters.}
Our research indicates that text prompts exert a significant influence on AIGC models. However, we observe that simple prompts, such as ``a teddy bear on a bench'' or ``a Picasso painting on the wall'', typically produce satisfactory results in the scenario of text-prompt inpainting. In contrast, longer and more complex prompts may yield impressive outcomes, but they tend to be less user-friendly.


\end{itemize}

\section{Experiment}
\begin{figure*}[t]
 \centering
  \begin{subfigure}{\linewidth}
   \centering
 \includegraphics[width=0.3\linewidth, trim=0cm 0cm 0cm 0cm]{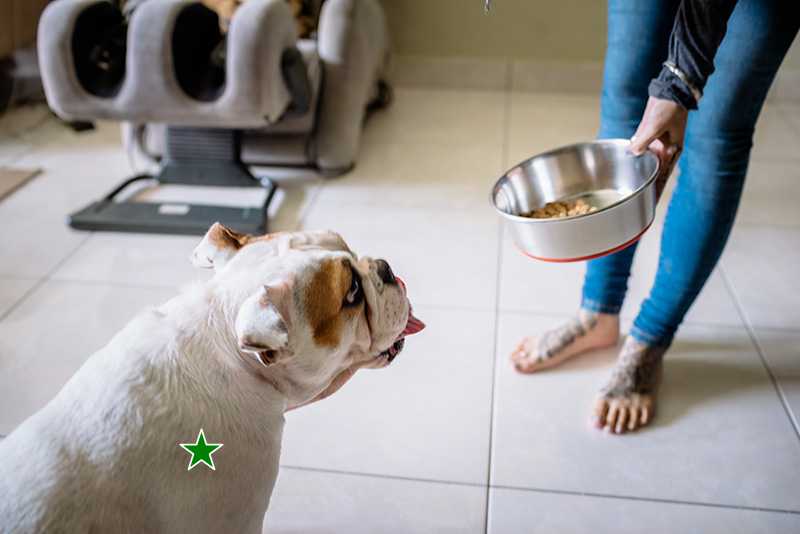}
  \includegraphics[width=0.3\linewidth, trim=0cm 0cm 0cm 0cm]{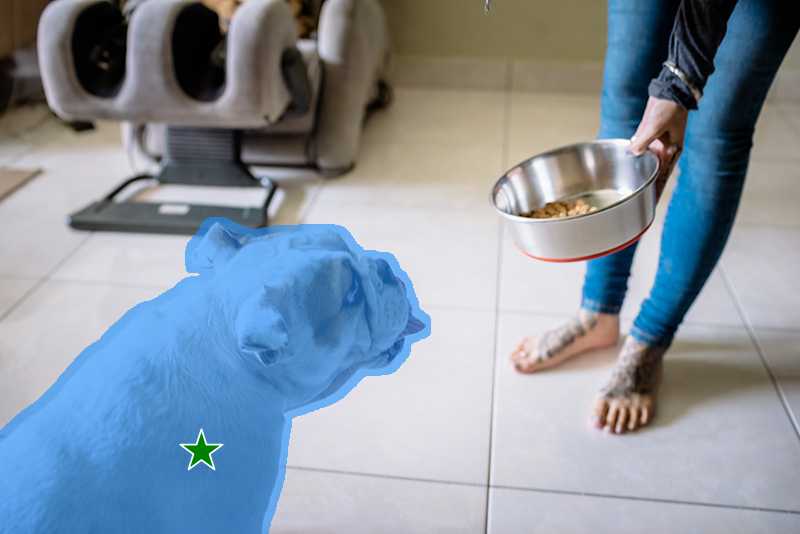}
   \includegraphics[width=0.3\linewidth, trim=0cm 0cm 0cm 0cm]{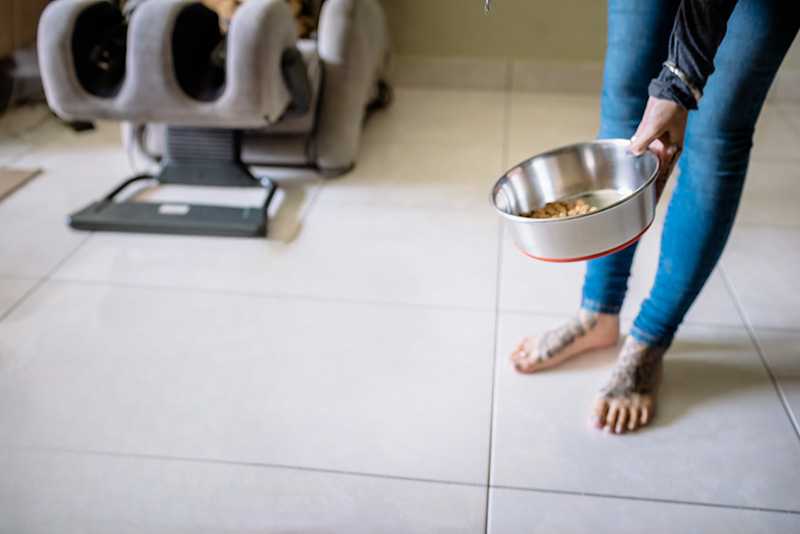}
 \end{subfigure}

  \begin{subfigure}{\linewidth}
   \centering
 \includegraphics[width=0.3\linewidth, trim=0cm 0cm 0cm 0cm]{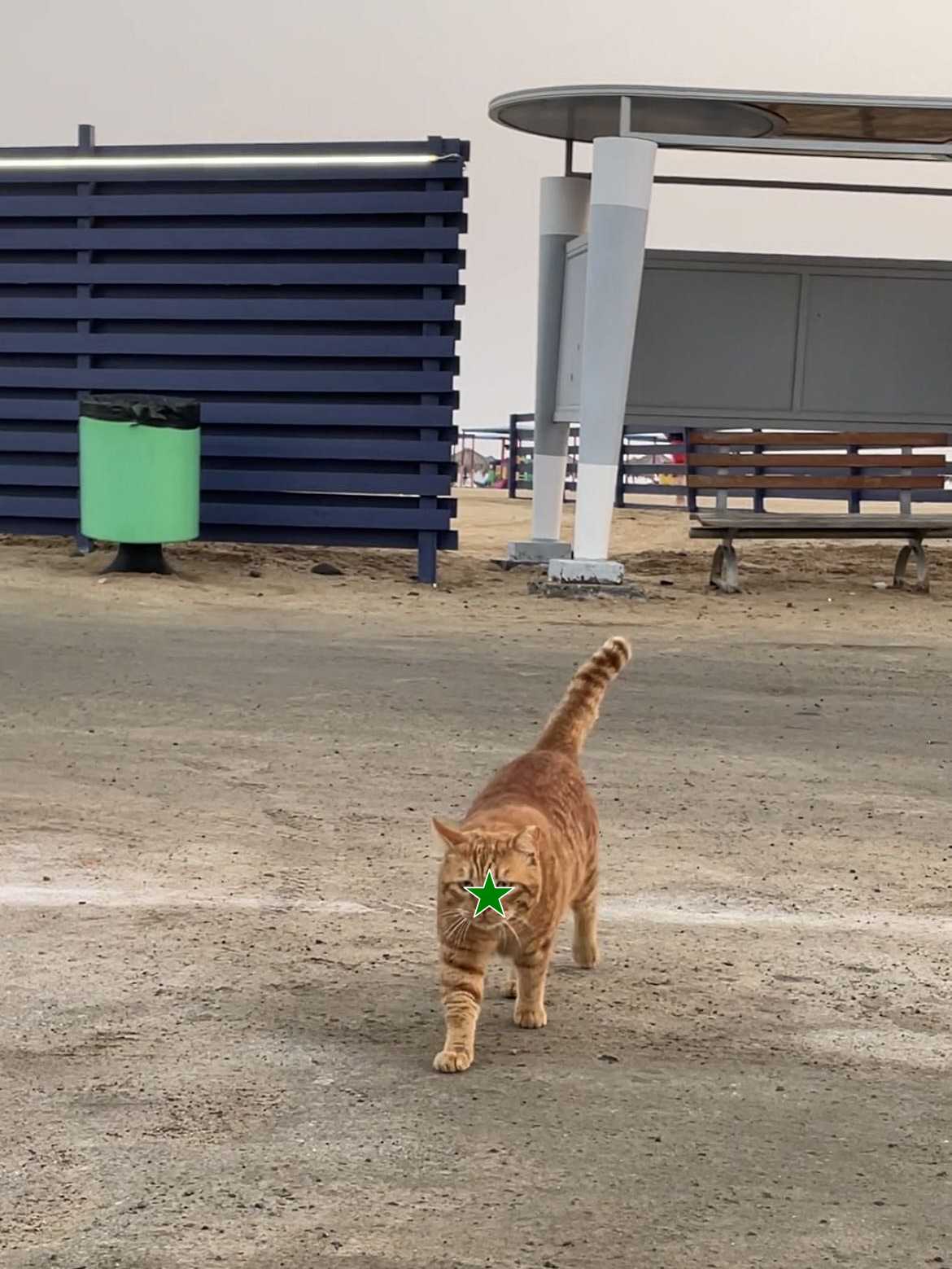}
  \includegraphics[width=0.3\linewidth, trim=0cm 0cm 0cm 0cm]{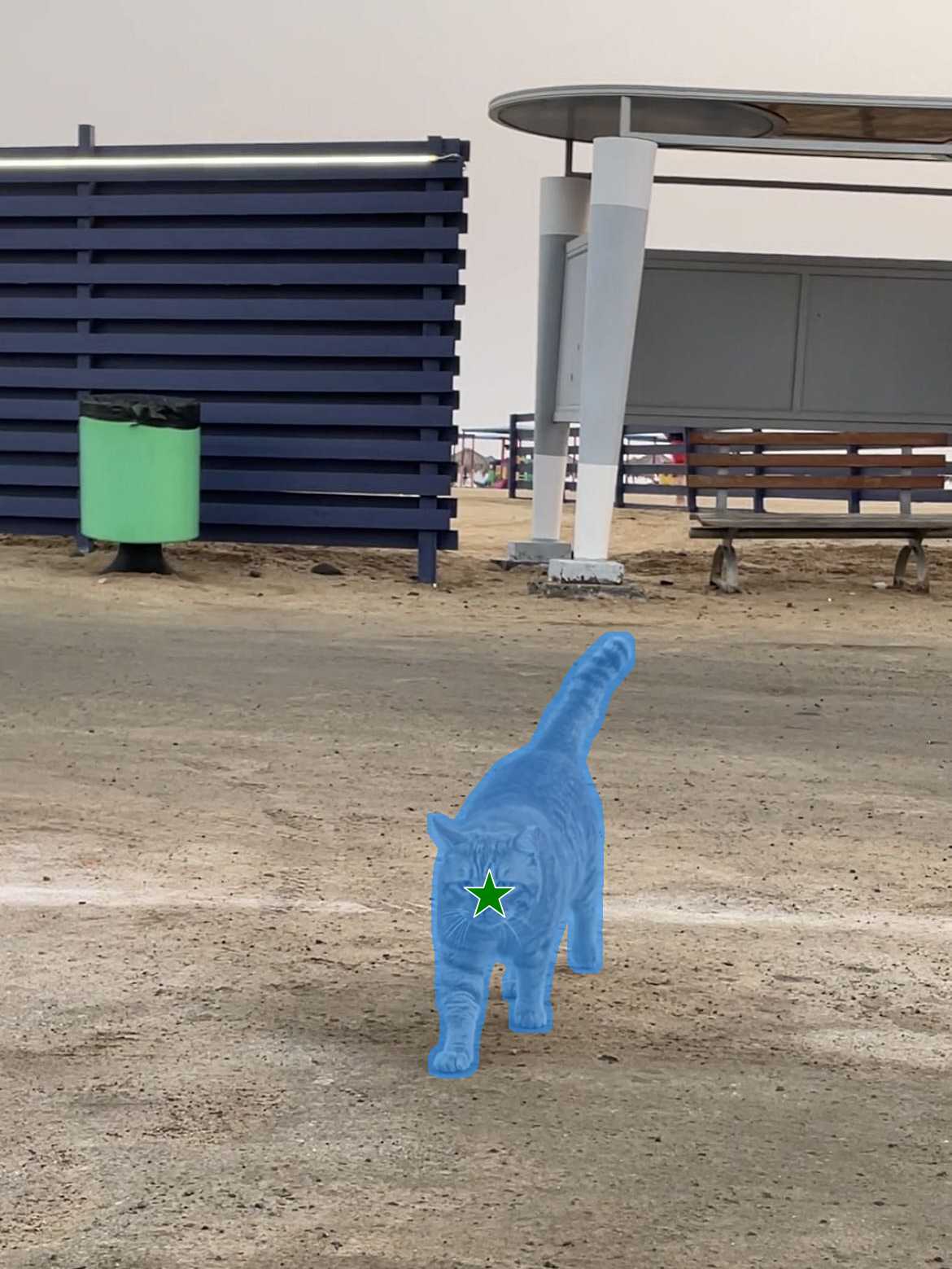}
   \includegraphics[width=0.3\linewidth, trim=0cm 0cm 0cm 0cm]{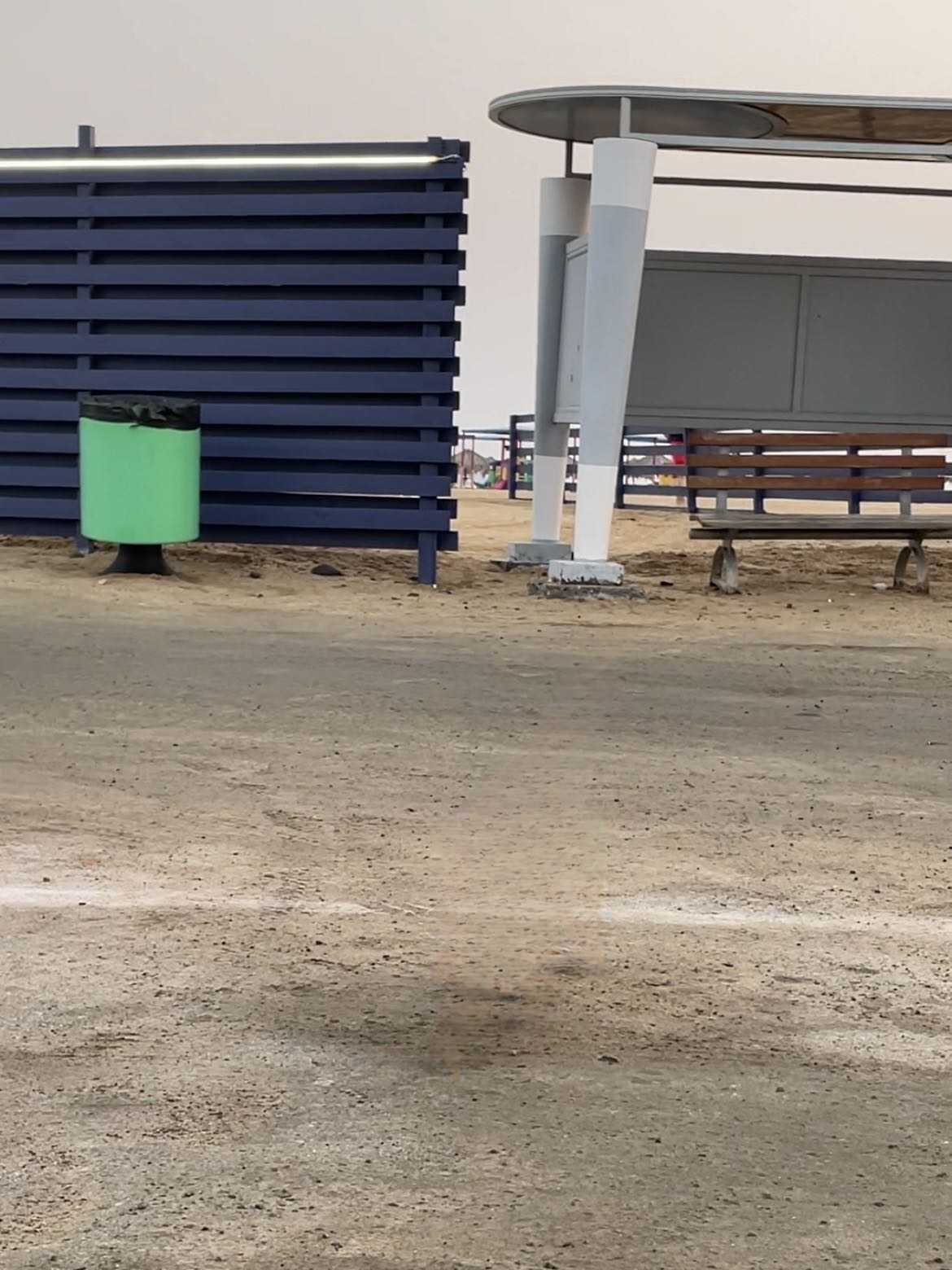}
 \end{subfigure}

   \begin{subfigure}{\linewidth}
    \centering
 \includegraphics[width=0.3\linewidth, trim=0cm 0cm 0cm 0cm]{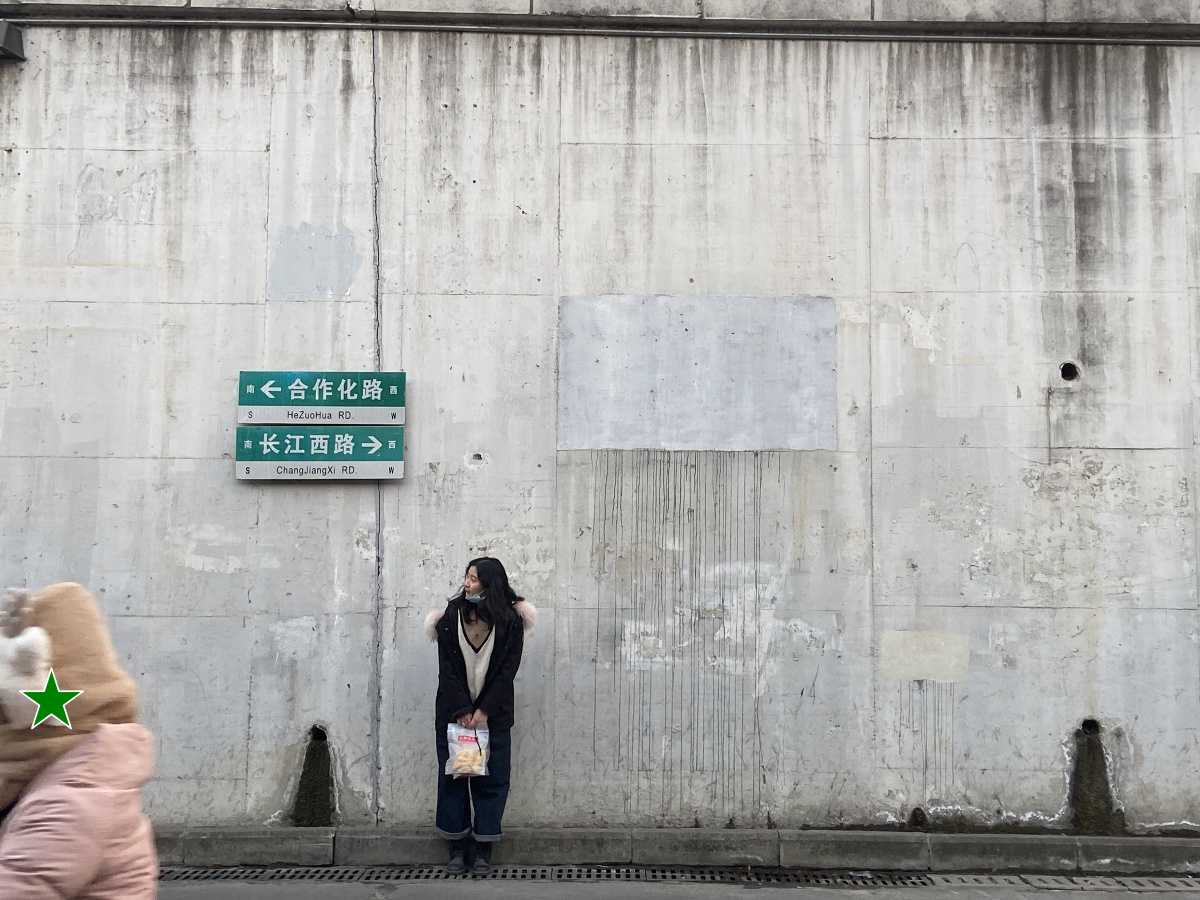}
  \includegraphics[width=0.3\linewidth, trim=0cm 0cm 0cm 0cm]{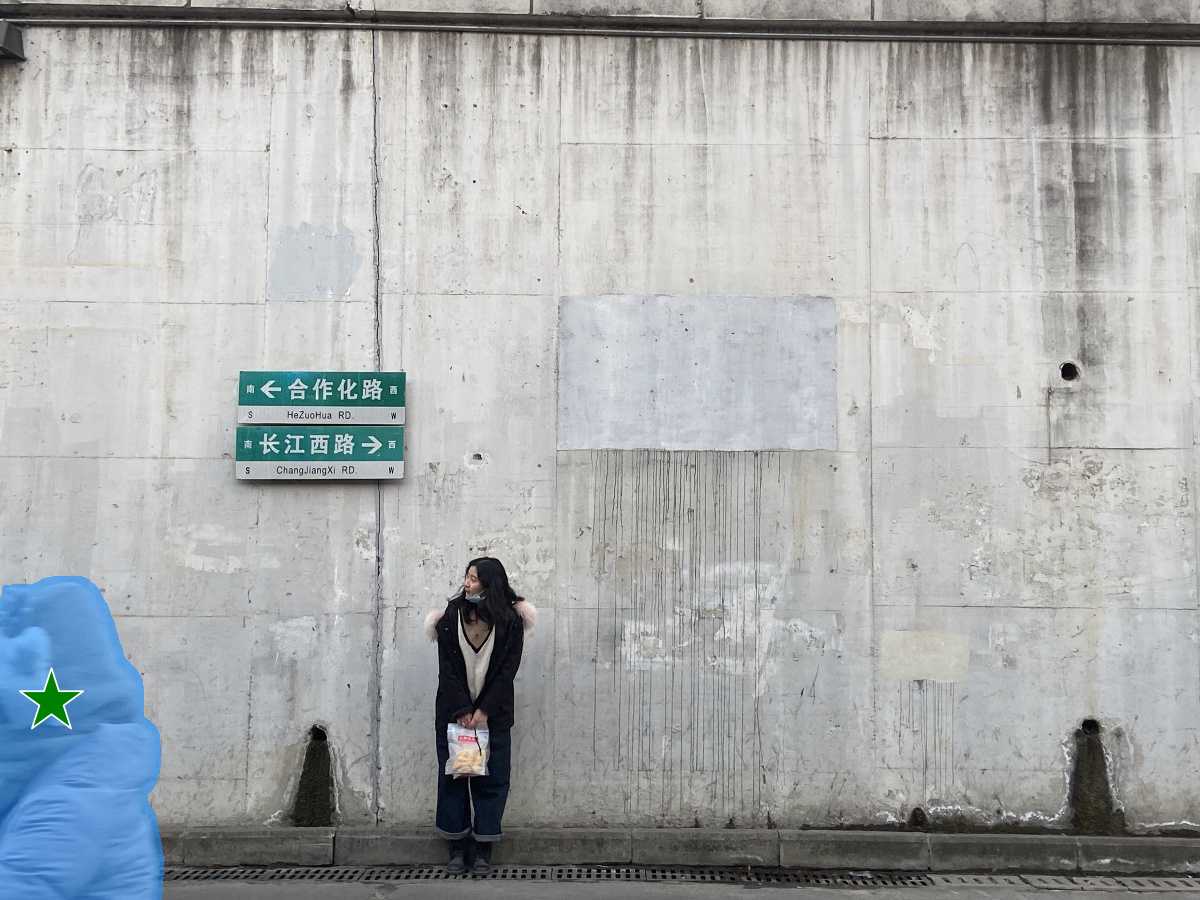}
   \includegraphics[width=0.3\linewidth, trim=0cm 0cm 0cm 0cm]{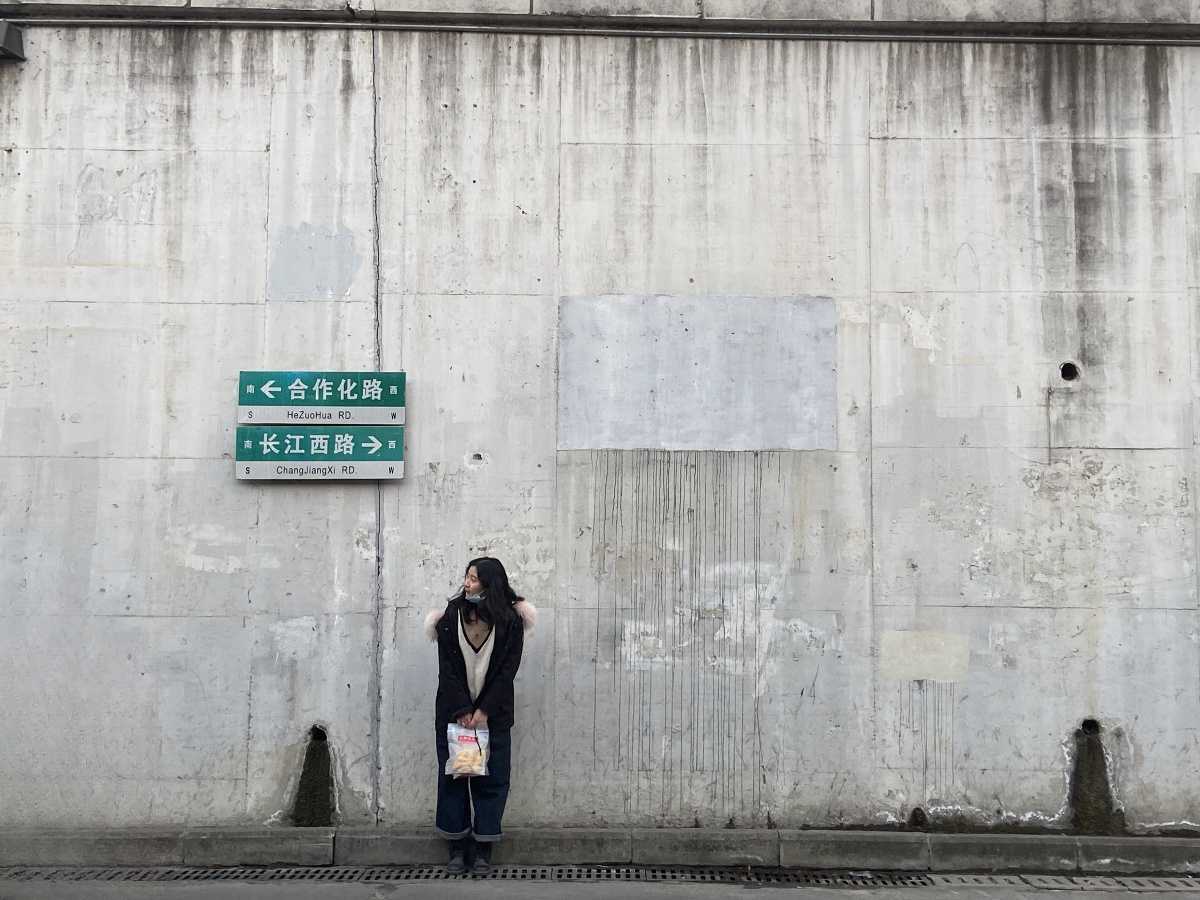}
 \end{subfigure}

   \begin{subfigure}{\linewidth}
    \centering
 \includegraphics[width=0.3\linewidth, trim=0cm 0cm 0cm 0cm]{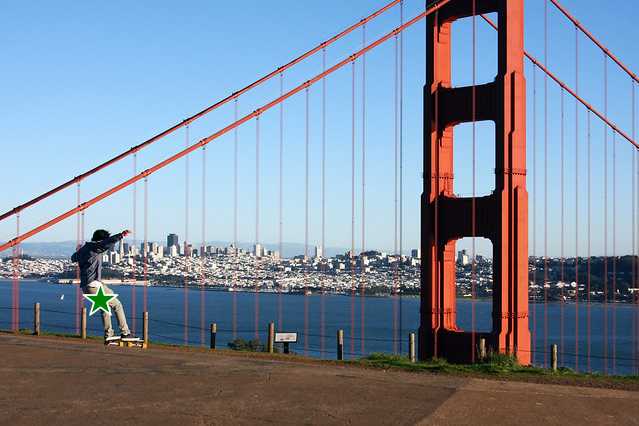}
  \includegraphics[width=0.3\linewidth, trim=0cm 0cm 0cm 0cm]{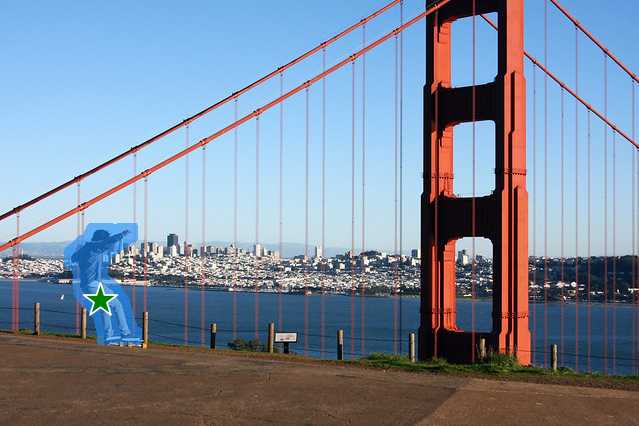}
   \includegraphics[width=0.3\linewidth, trim=0cm 0cm 0cm 0cm]{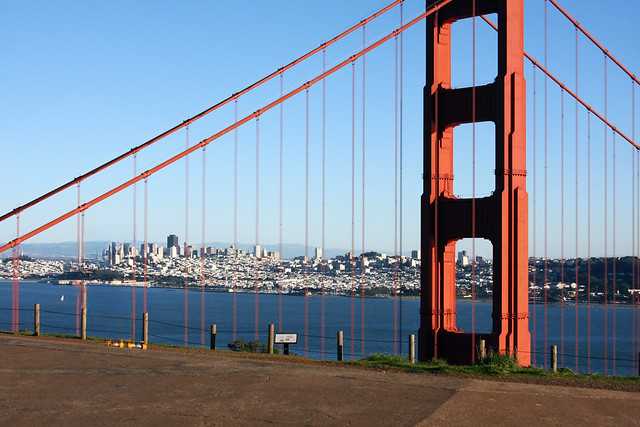}
 \end{subfigure}


\begin{subfigure}{\linewidth}
 \centering
 \includegraphics[width=0.3\linewidth, trim=0cm 0cm 0cm 0cm]{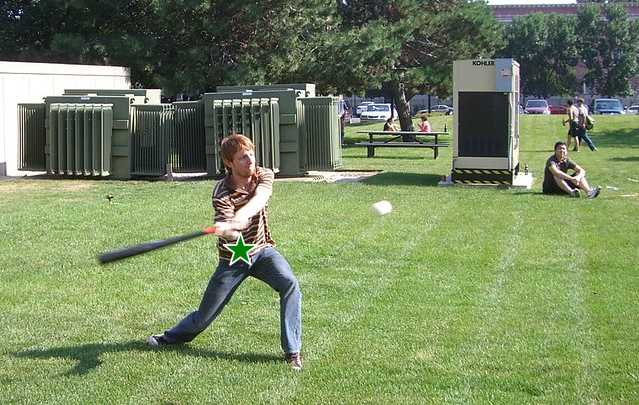}
  \includegraphics[width=0.3\linewidth, trim=0cm 0cm 0cm 0cm]{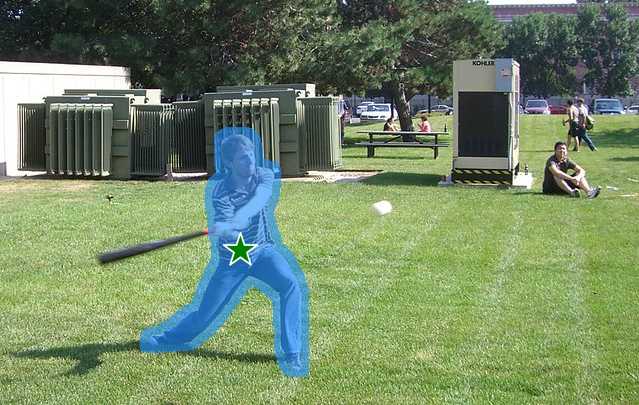}
   \includegraphics[width=0.3\linewidth, trim=0cm 0cm 0cm 0cm]{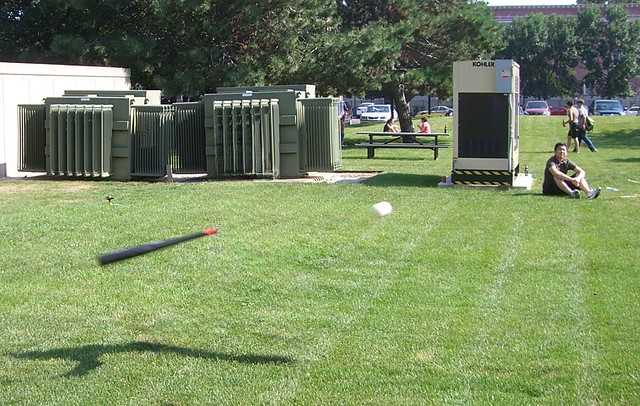}
 \end{subfigure}

\vspace{-2mm}
\caption{Visualization results of Remove Anything. }
\label{fig:demo-remove-anything}
\vspace{-2mm}
\end{figure*}

\begin{figure*}[t]
 \centering
  \begin{subfigure}{\linewidth}
   \centering
 \includegraphics[width=0.3\linewidth, trim=0cm 0cm 0cm 0cm]{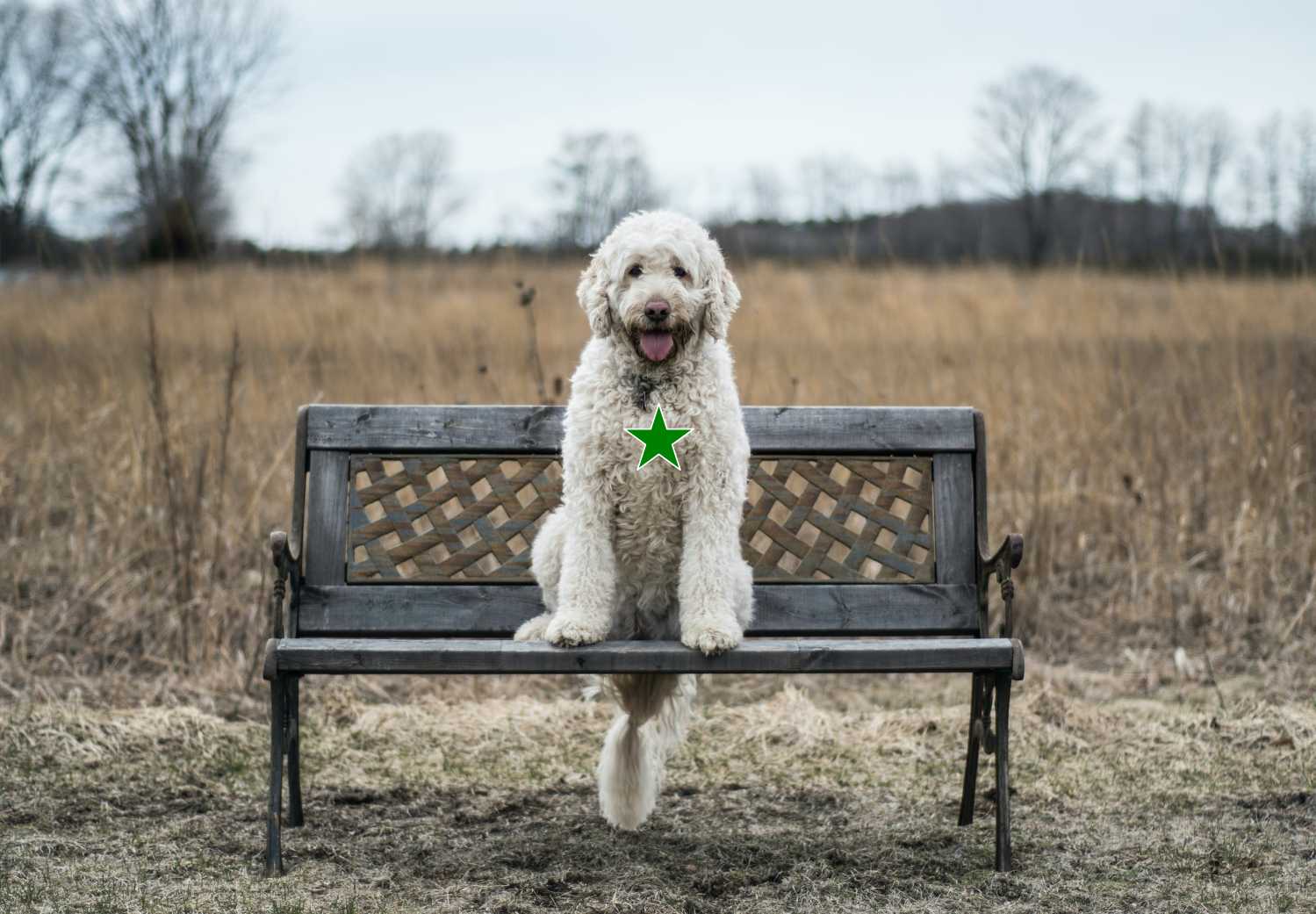}
  \includegraphics[width=0.3\linewidth, trim=0cm 0cm 0cm 0cm]{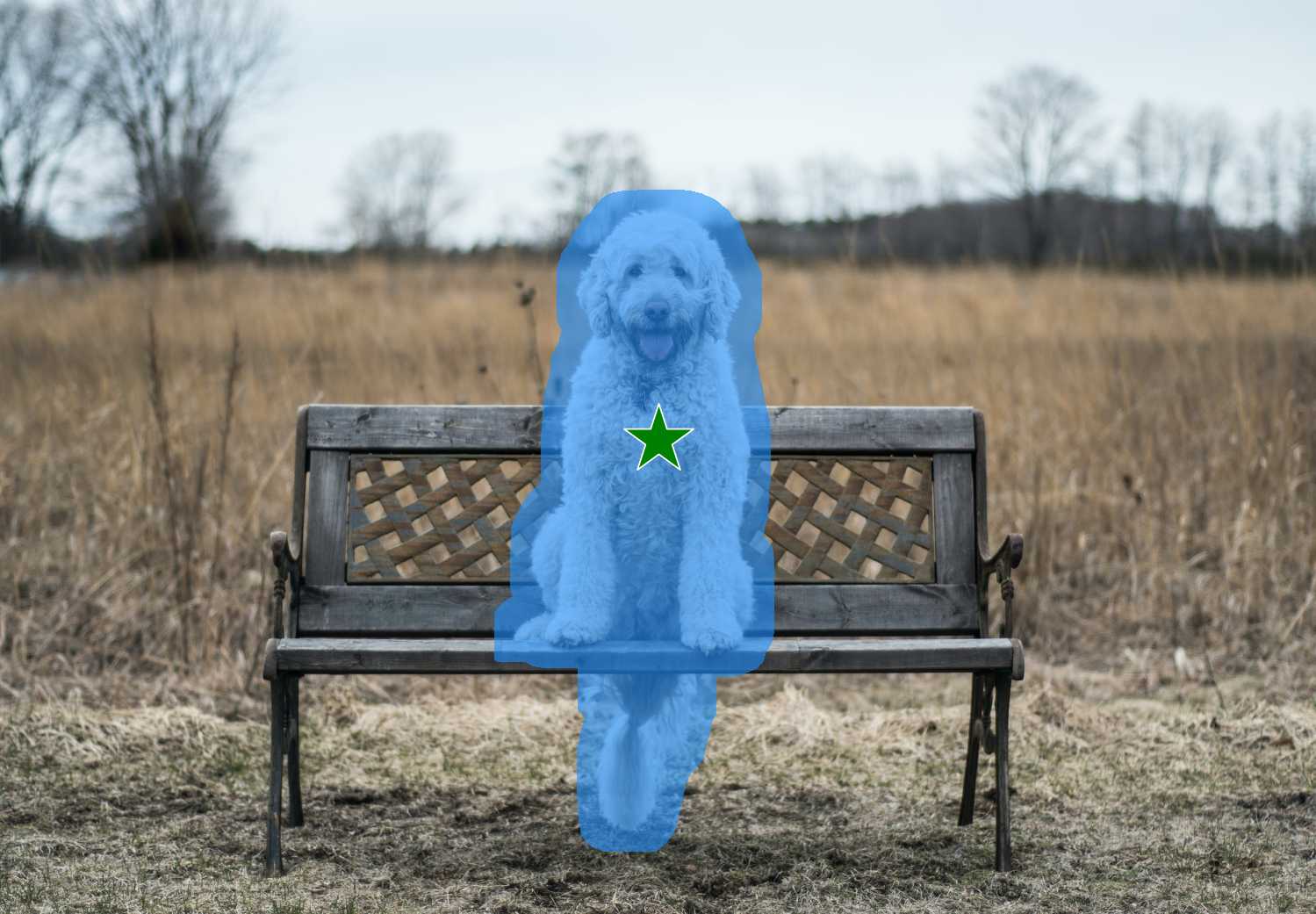}
   \includegraphics[width=0.3\linewidth, trim=0cm 0cm 0cm 0cm]{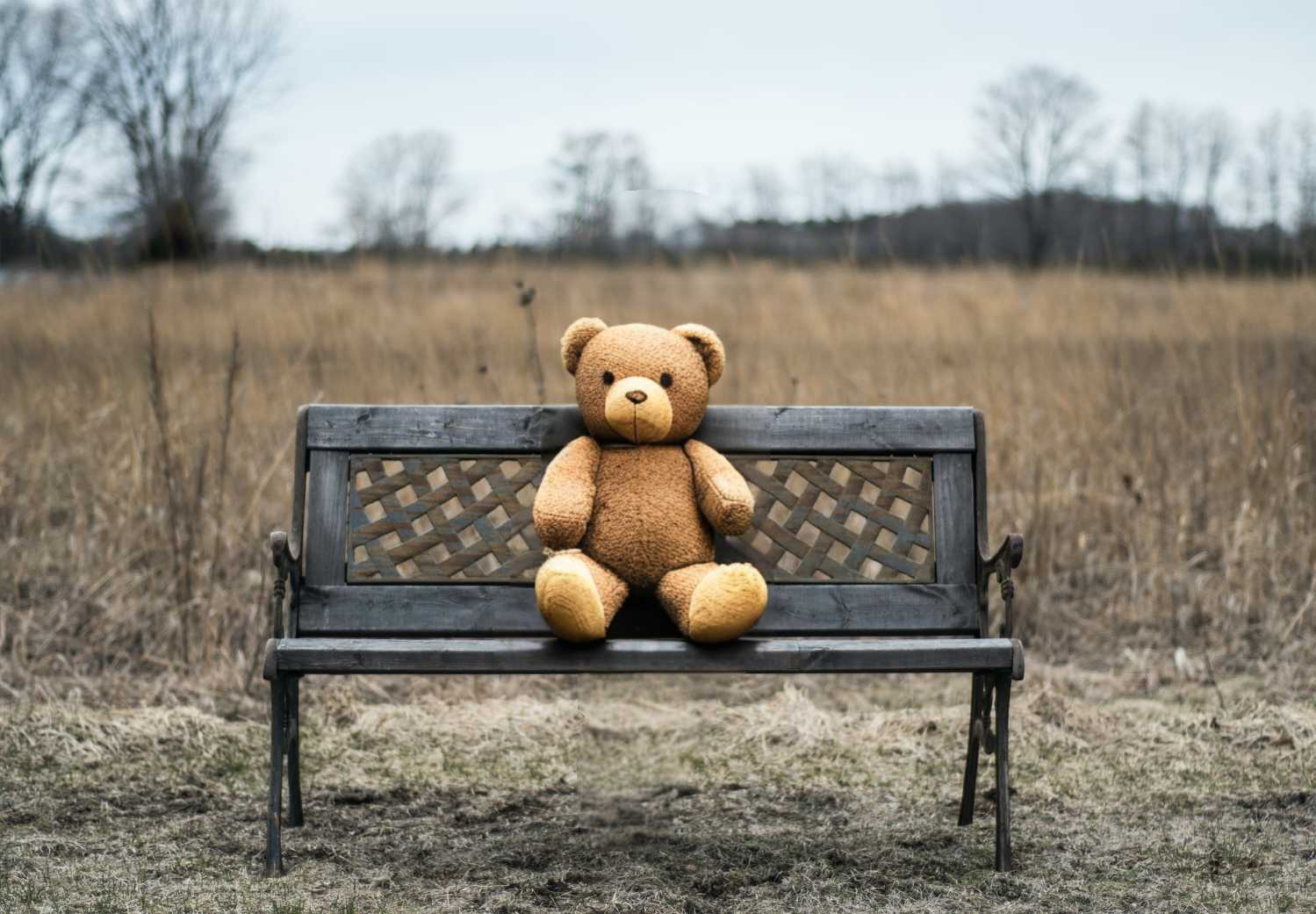}
   \vspace{-2mm}
\caption{Text prompt: a teddy bear on a bench}
 \end{subfigure}

\begin{subfigure}{\linewidth}
 \centering
 \includegraphics[width=0.3\linewidth, trim=0cm 0cm 0cm 0cm]{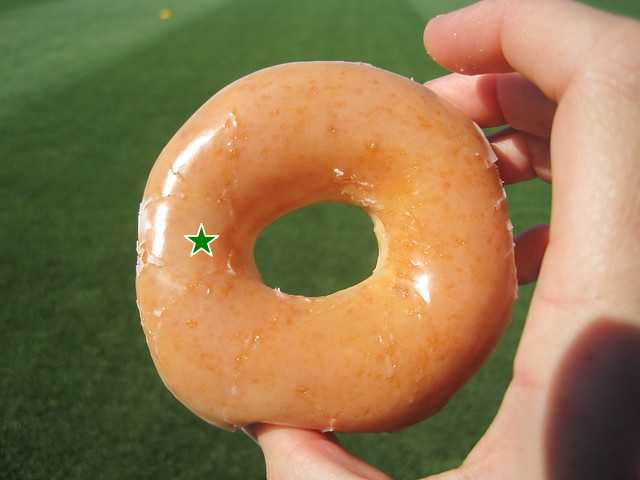}
  \includegraphics[width=0.3\linewidth, trim=0cm 0cm 0cm 0cm]{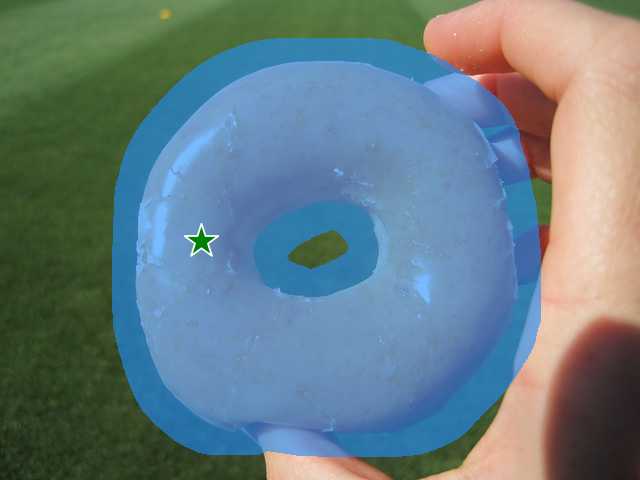}
   \includegraphics[width=0.3\linewidth, trim=0cm 0cm 0cm 0cm]{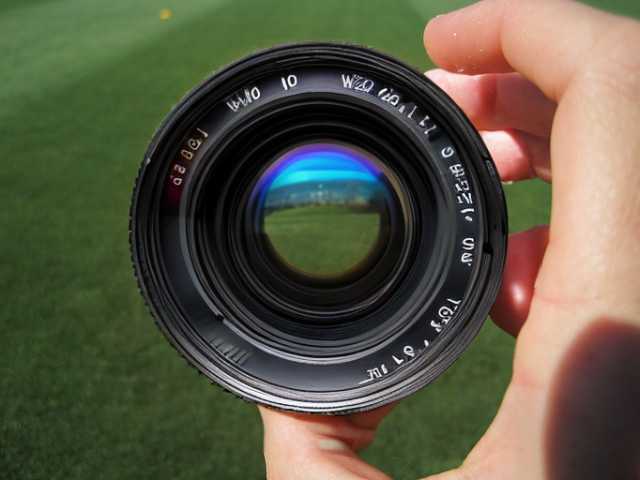}
   \vspace{-2mm}
\caption{Text prompt: a camera lens in the hand}
 \end{subfigure}

\begin{subfigure}{\linewidth}
 \centering
 \includegraphics[width=0.3\linewidth, trim=0cm 0cm 0cm 0cm]{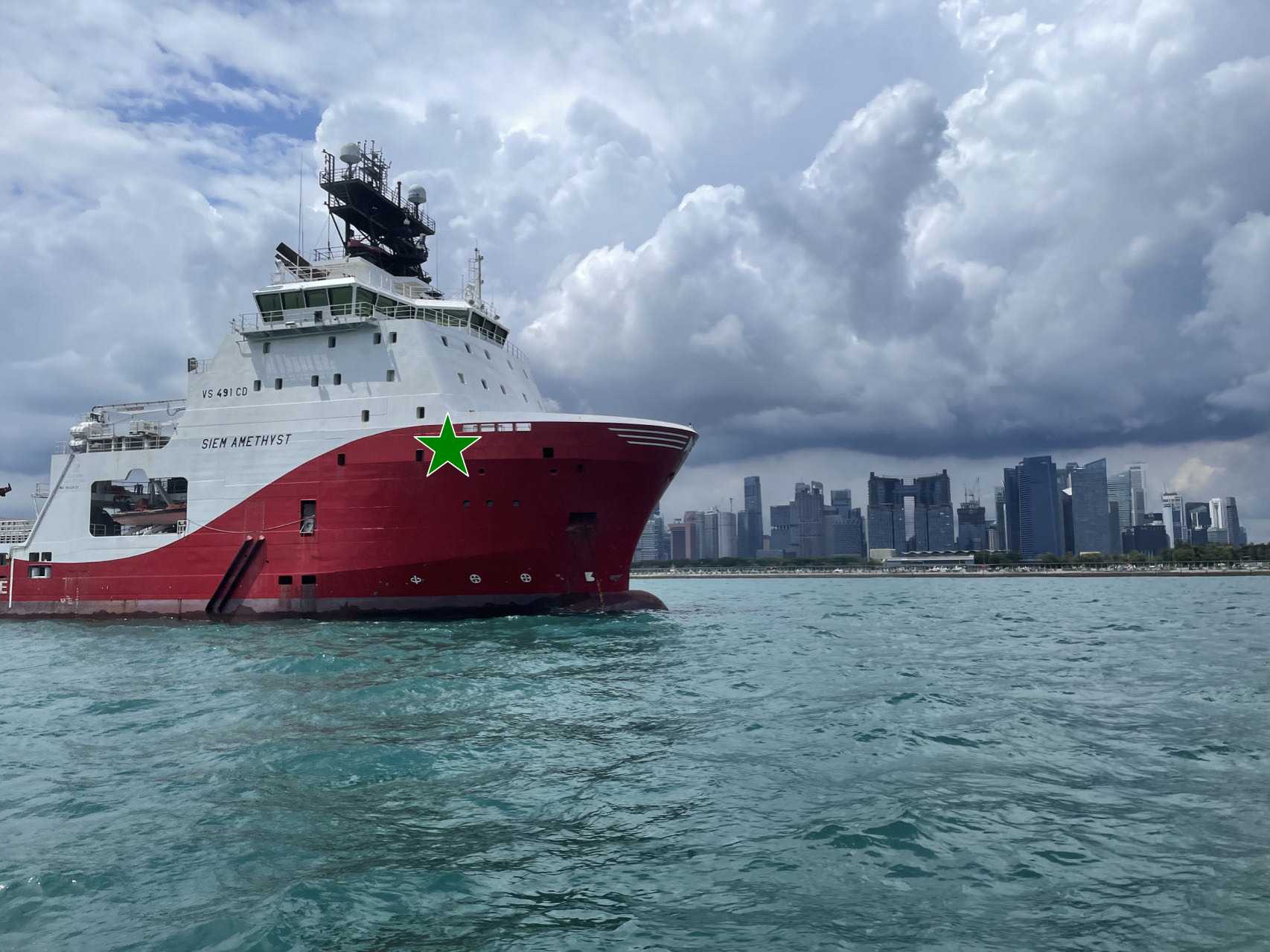}
  \includegraphics[width=0.3\linewidth, trim=0cm 0cm 0cm 0cm]{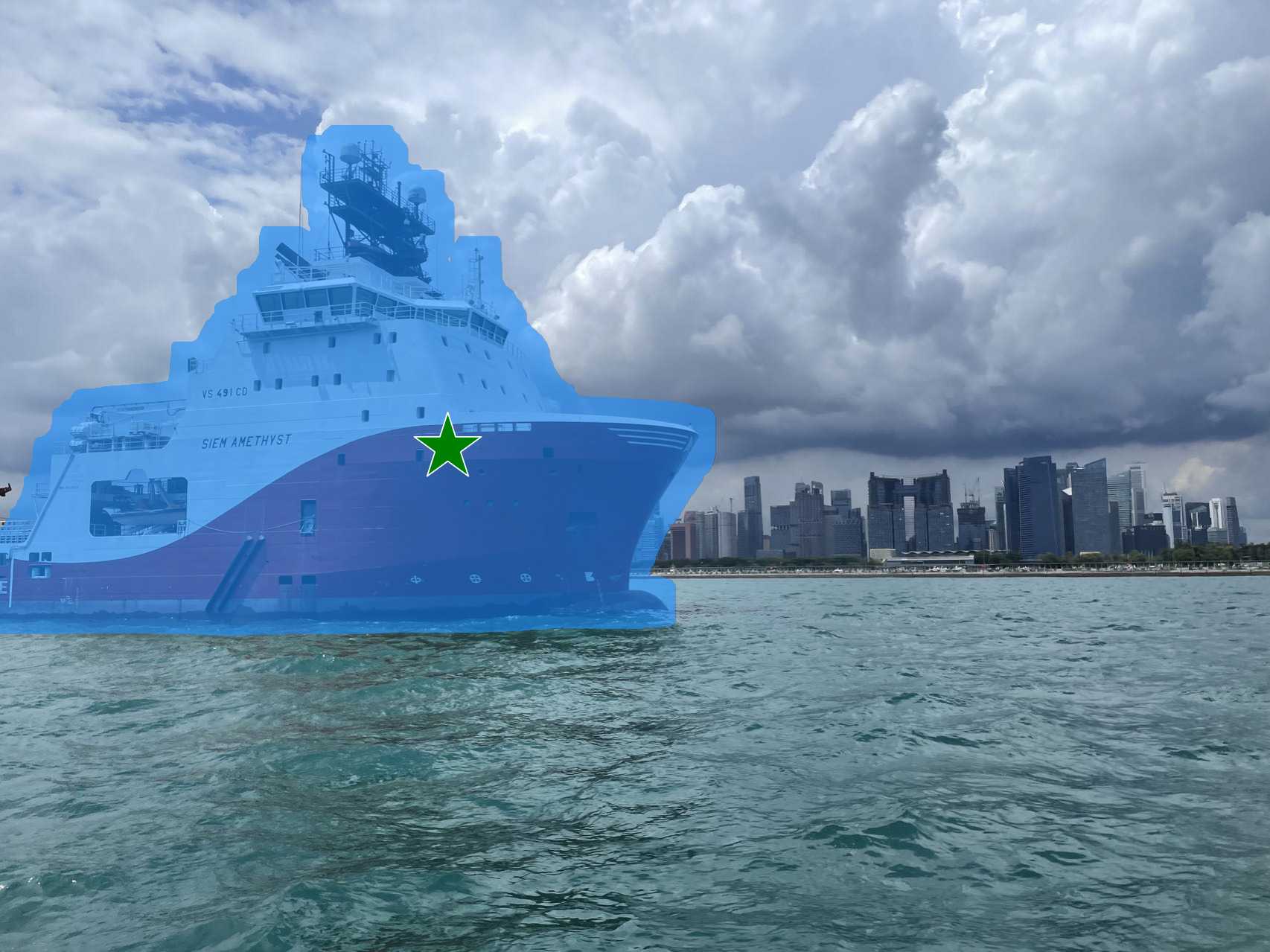}
   \includegraphics[width=0.3\linewidth, trim=0cm 0cm 0cm 0cm]{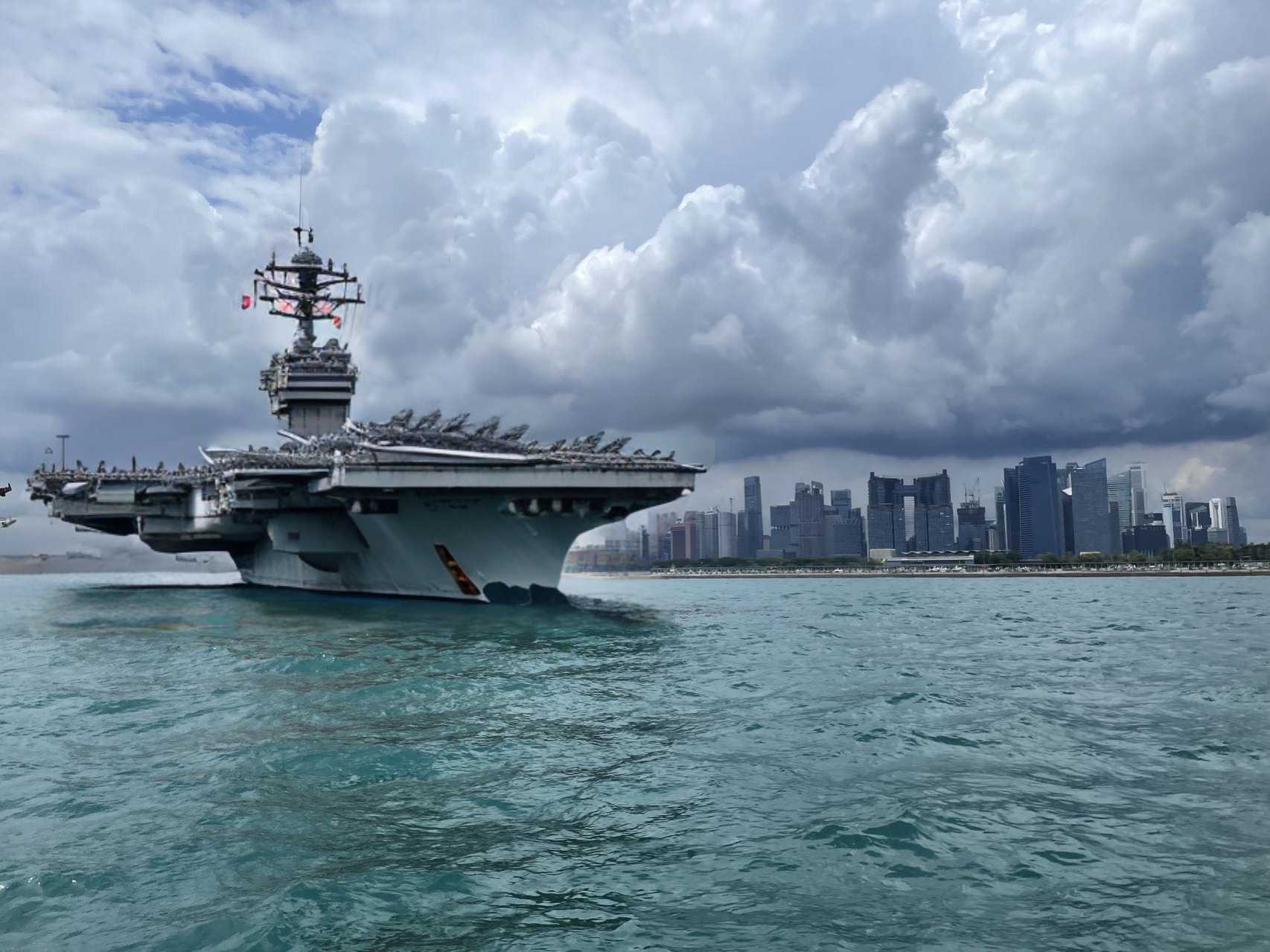}
   \vspace{-2mm}
\caption{Text prompt: an aircraft carrier on the sea}
 \end{subfigure}

\begin{subfigure}{\linewidth}
 \centering
 \includegraphics[width=0.3\linewidth, trim=0cm 0cm 0cm 0cm]{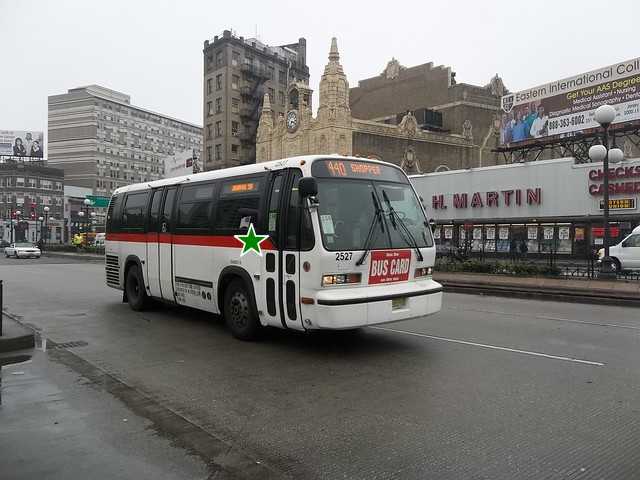}
  \includegraphics[width=0.3\linewidth, trim=0cm 0cm 0cm 0cm]{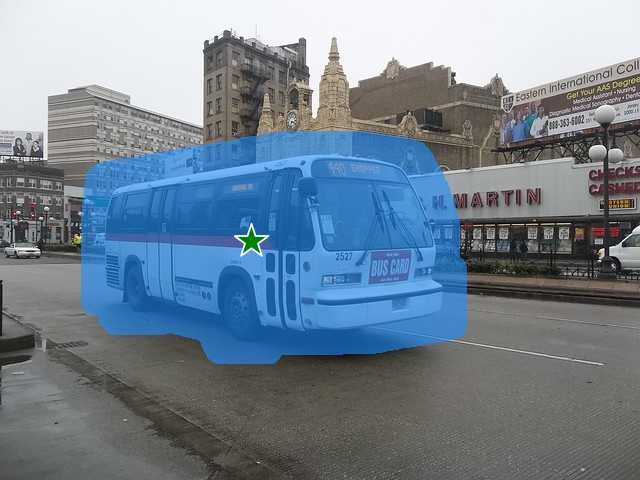}
   \includegraphics[width=0.3\linewidth, trim=0cm 0cm 0cm 0cm]{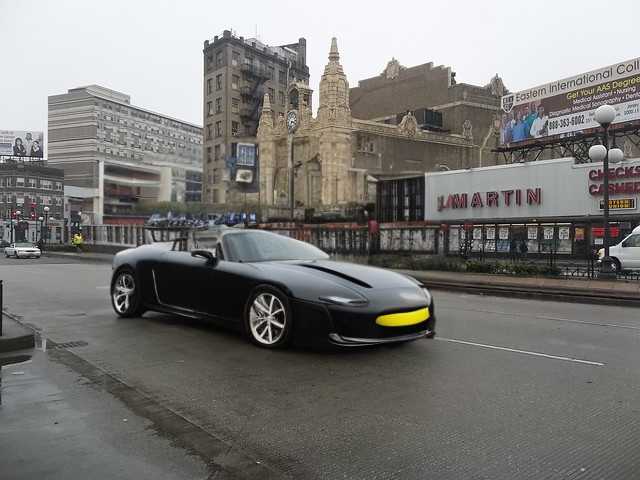}
   \vspace{-2mm}
\caption{Text prompt: a sports car on a road}

 \end{subfigure}

 

 \begin{subfigure}{\linewidth}
  \centering
 \includegraphics[width=0.3\linewidth, trim=0cm 0cm 0cm 0cm]{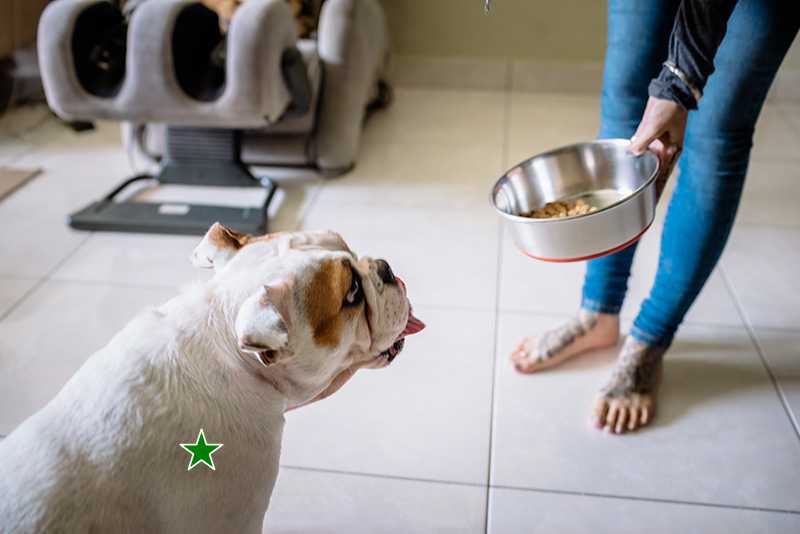}
  \includegraphics[width=0.3\linewidth, trim=0cm 0cm 0cm 0cm]{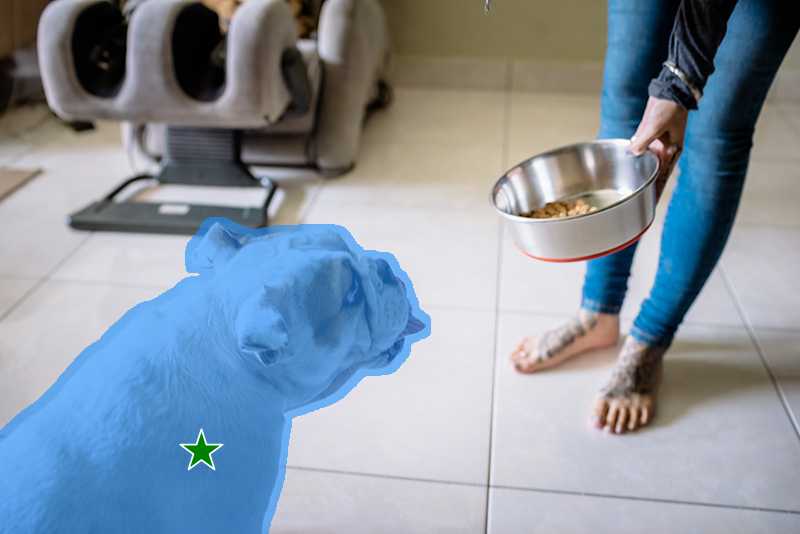}
   \includegraphics[width=0.3\linewidth, trim=0cm 0cm 0cm 0cm]{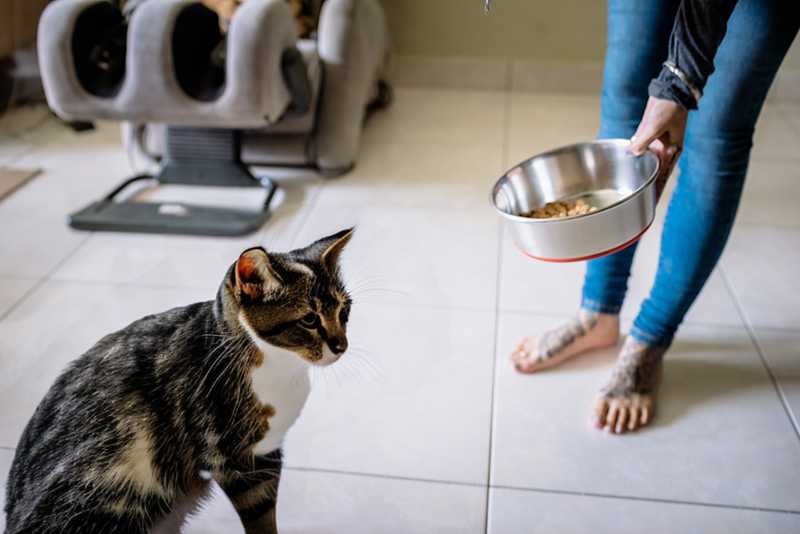}
\vspace{-2mm}
\caption{Text prompt: a cat, waiting for food}
 \end{subfigure}

\vspace{-2mm}
\caption{Visualization results of Fill Anything. }
\label{fig:demo-fill-anything}
\vspace{-2mm}
\end{figure*}

\begin{figure*}[t]
 \centering
  \begin{subfigure}{\linewidth}
   \centering
 \includegraphics[width=0.3\linewidth, trim=0cm 0cm 0cm 0cm]{./figures/replace-anything/000000000724/with_points.jpg}
  \includegraphics[width=0.3\linewidth, trim=0cm 0cm 0cm 0cm]{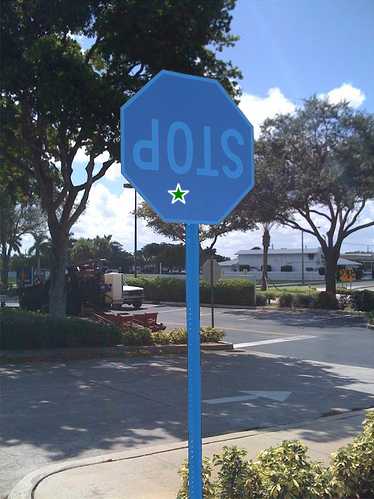}
   \includegraphics[width=0.3\linewidth, trim=0cm 0cm 0cm 0cm]{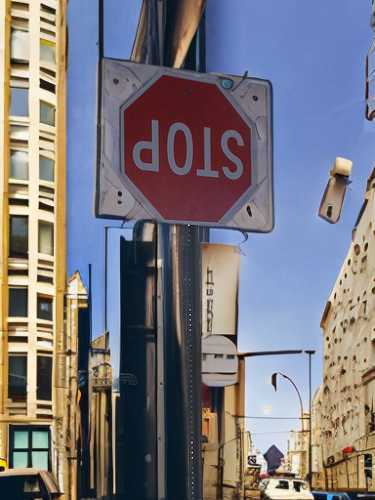}
   \vspace{-2mm}
\caption{Text prompt: crossroad in the city}
 \end{subfigure}

  \begin{subfigure}{\linewidth}
   \centering
 \includegraphics[width=0.3\linewidth, trim=0cm 0cm 0cm 0cm]{./figures/replace-anything/000000029675/with_points.jpg}
  \includegraphics[width=0.3\linewidth, trim=0cm 0cm 0cm 0cm]{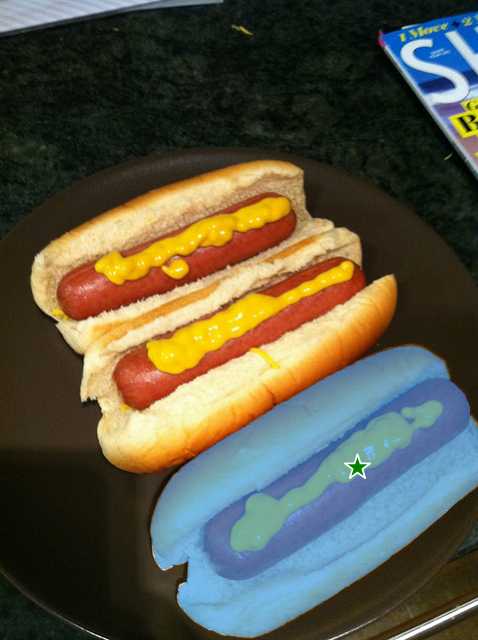}
   \includegraphics[width=0.3\linewidth, trim=0cm 0cm 0cm 0cm]{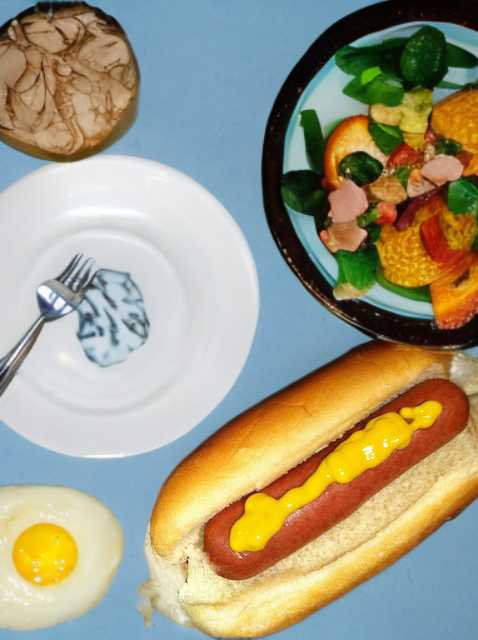}
   \vspace{-2mm}
 \caption{Text prompt: breakfast}
 \end{subfigure}

   \begin{subfigure}{\linewidth}
    \centering
 \includegraphics[width=0.3\linewidth, trim=0cm 0cm 0cm 0cm]{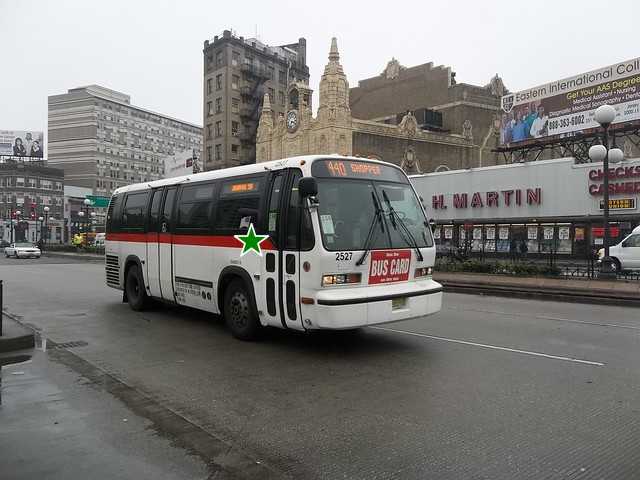}
  \includegraphics[width=0.3\linewidth, trim=0cm 0cm 0cm 0cm]{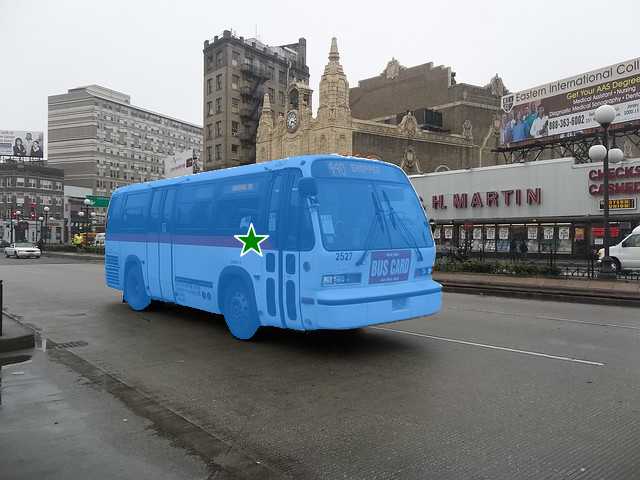}
   \includegraphics[width=0.3\linewidth, trim=0cm 0cm 0cm 0cm]{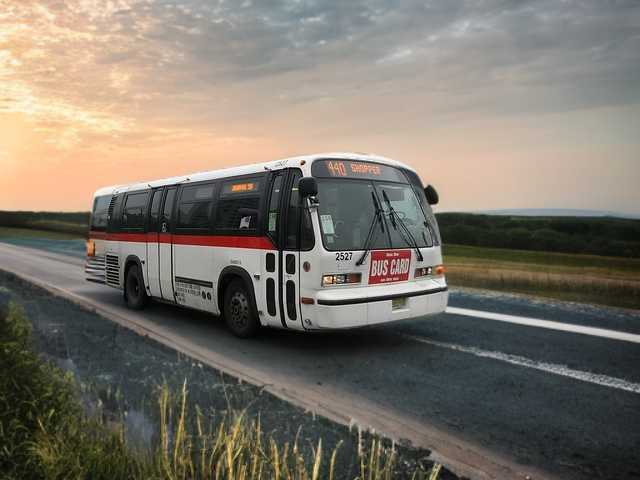}
   \vspace{-2mm}
 \caption{Text prompt: a bus, on the center of a country road, summer evening}
 \end{subfigure}

   \begin{subfigure}{\linewidth}
    \centering
 \includegraphics[width=0.3\linewidth, trim=0cm 0cm 0cm 0cm]{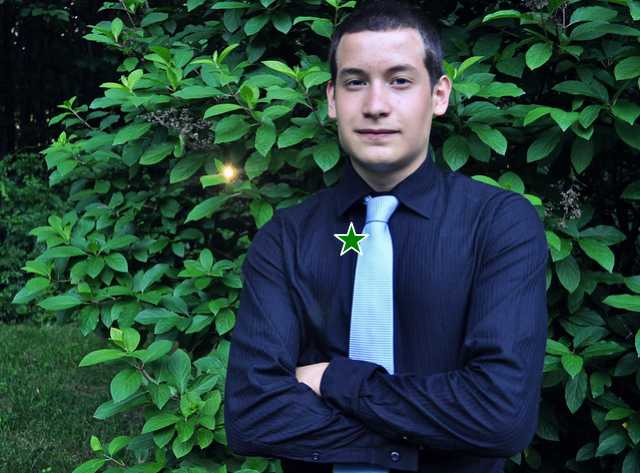}
  \includegraphics[width=0.3\linewidth, trim=0cm 0cm 0cm 0cm]{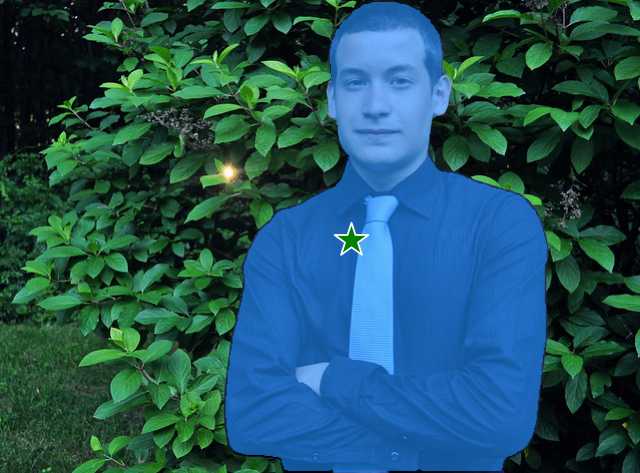}
   \includegraphics[width=0.3\linewidth, trim=0cm 0cm 0cm 0cm]{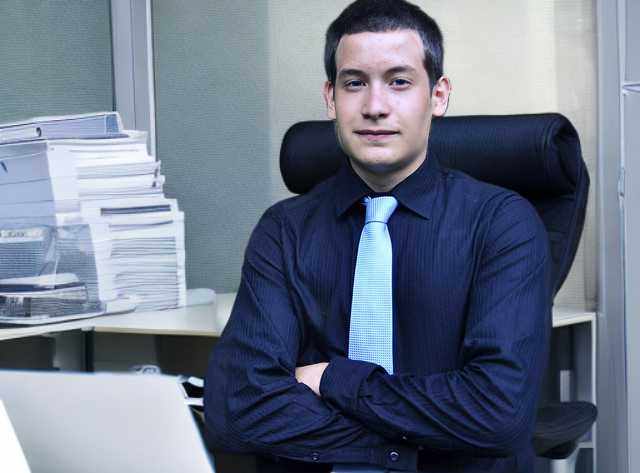}
   \vspace{-2mm}
 \caption{Text prompt: a man in office}
 \end{subfigure}

\vspace{-2mm}
\caption{Visualization results of Replace Anything. }
\label{fig:demo-replace-anything}
\vspace{-2mm}
\end{figure*}

We evaluate Remove Anything, Fill Anything and Replace Anything in our Inpaint Anything in three cases, \ieno, removing objects, filling objects and replacing background, respectively. We collect test images from COCO dataset \cite{lin2014microsoft}, LaMa test set \cite{suvorov2022resolution} and photos taken by our phones.
The results are in Figure \ref{fig:demo-remove-anything}, \ref{fig:demo-fill-anything} and \ref{fig:demo-replace-anything}. The experimental results indicate that the proposed Inpaint Anything is both general and robust, effectively inpainting images with diverse content, resolutions, and aspect ratios.




\section{Conclusion}

Inpaint Anything (IA) is a versatile tool that combines the capabilities of \emph{Remove Anything, Fill Anything, and Replace Anything}. Based on the vision foundation models of Segmentation Anything, SOTA inpainter and AIGC models, IA enables to realize mask-free image inpainting, and also supports the user-friendly operation of ``click for removing, and prompt for filling''. Besides, IA can handle more various and high-quality input images, with any aspect ratio and 2K resolution. We build such interesting project to demonstrate a strong power of fully exploiting the existing LARGE AI models, and reveal the potential of ``Composable AI''. We are also very willing to help everyone share and promote new projects based on our Inpaint Anything (IA). In the future, we will further develop our Inpaint Anything (IA) to support more practical functions, like fine-grained image matting, editing, etc., and apply it to more realistic applications.

{\small
\bibliographystyle{ieee_fullname}
\bibliography{egbib}

\begin{thebibliography}{10}\itemsep=-1pt

\bibitem{chi2020fast}
Lu Chi, Borui Jiang, and Yadong Mu.
\newblock Fast fourier convolution.
\newblock {\em Advances in Neural Information Processing Systems},
  33:4479--4488, 2020.

\bibitem{criminisi2003object}
Antonio Criminisi, Patrick Perez, and Kentaro Toyama.
\newblock Object removal by exemplar-based inpainting.
\newblock In {\em 2003 IEEE Computer Society Conference on Computer Vision and
  Pattern Recognition, 2003. Proceedings.}, volume~2, pages II--II. IEEE, 2003.

\bibitem{criminisi2004region}
Antonio Criminisi, Patrick P{\'e}rez, and Kentaro Toyama.
\newblock Region filling and object removal by exemplar-based image inpainting.
\newblock {\em IEEE Transactions on Image Processing}, 13(9):1200--1212, 2004.

\bibitem{dong2022incremental}
Qiaole Dong, Chenjie Cao, and Yanwei Fu.
\newblock Incremental transformer structure enhanced image inpainting with
  masking positional encoding.
\newblock In {\em Proceedings of the IEEE/CVF Conference on Computer Vision and
  Pattern Recognition}, pages 11358--11368, 2022.

\bibitem{elharrouss2020image}
Omar Elharrouss, Noor Almaadeed, Somaya Al-Maadeed, and Younes Akbari.
\newblock Image inpainting: A review.
\newblock {\em Neural Processing Letters}, 51:2007--2028, 2020.

\bibitem{johnson2016perceptual}
Justin Johnson, Alexandre Alahi, and Li Fei-Fei.
\newblock Perceptual losses for real-time style transfer and super-resolution.
\newblock In {\em Computer Vision--ECCV 2016: 14th European Conference,
  Amsterdam, The Netherlands, October 11-14, 2016, Proceedings, Part II 14},
  pages 694--711. Springer, 2016.

\bibitem{kirillov2023segment}
Alexander Kirillov, Eric Mintun, Nikhila Ravi, Hanzi Mao, Chloe Rolland, Laura
  Gustafson, Tete Xiao, Spencer Whitehead, Alexander~C Berg, Wan-Yen Lo, et~al.
\newblock Segment anything.
\newblock {\em arXiv preprint arXiv:2304.02643}, 2023.

\bibitem{li2022mat}
Wenbo Li, Zhe Lin, Kun Zhou, Lu Qi, Yi Wang, and Jiaya Jia.
\newblock Mat: Mask-aware transformer for large hole image inpainting.
\newblock In {\em Proceedings of the IEEE/CVF conference on computer vision and
  pattern recognition}, pages 10758--10768, 2022.

\bibitem{lin2014microsoft}
Tsung-Yi Lin, Michael Maire, Serge Belongie, James Hays, Pietro Perona, Deva
  Ramanan, Piotr Doll{\'a}r, and C~Lawrence Zitnick.
\newblock Microsoft coco: Common objects in context.
\newblock In {\em Computer Vision--ECCV 2014: 13th European Conference, Zurich,
  Switzerland, September 6-12, 2014, Proceedings, Part V 13}, pages 740--755.
  Springer, 2014.

\bibitem{lugmayr2022repaint}
Andreas Lugmayr, Martin Danelljan, Andres Romero, Fisher Yu, Radu Timofte, and
  Luc Van~Gool.
\newblock Repaint: Inpainting using denoising diffusion probabilistic models.
\newblock In {\em Proceedings of the IEEE/CVF Conference on Computer Vision and
  Pattern Recognition}, pages 11461--11471, 2022.

\bibitem{rombach2022high}
Robin Rombach, Andreas Blattmann, Dominik Lorenz, Patrick Esser, and Bj{\"o}rn
  Ommer.
\newblock High-resolution image synthesis with latent diffusion models.
\newblock In {\em Proceedings of the IEEE/CVF Conference on Computer Vision and
  Pattern Recognition}, pages 10684--10695, 2022.

\bibitem{saharia2022photorealistic}
Chitwan Saharia, William Chan, Saurabh Saxena, Lala Li, Jay Whang, Emily~L
  Denton, Kamyar Ghasemipour, Raphael Gontijo~Lopes, Burcu Karagol~Ayan, Tim
  Salimans, et~al.
\newblock Photorealistic text-to-image diffusion models with deep language
  understanding.
\newblock {\em Advances in Neural Information Processing Systems},
  35:36479--36494, 2022.

\bibitem{suvorov2022resolution}
Roman Suvorov, Elizaveta Logacheva, Anton Mashikhin, Anastasia Remizova,
  Arsenii Ashukha, Aleksei Silvestrov, Naejin Kong, Harshith Goka, Kiwoong
  Park, and Victor Lempitsky.
\newblock Resolution-robust large mask inpainting with fourier convolutions.
\newblock In {\em Proceedings of the IEEE/CVF winter conference on applications
  of computer vision}, pages 2149--2159, 2022.

\end{thebibliography}
}

\end{document}